\documentclass[letterpaper]{article} % DO NOT CHANGE THIS
\usepackage{aaai24}  % DO NOT CHANGE THIS
\usepackage{times}  % DO NOT CHANGE THIS
\usepackage{helvet}  % DO NOT CHANGE THIS
\usepackage{courier}  % DO NOT CHANGE THIS
\usepackage[hyphens]{url}  % DO NOT CHANGE THIS
\usepackage{graphicx} % DO NOT CHANGE THIS
\usepackage{natbib}  % DO NOT CHANGE THIS AND DO NOT ADD ANY OPTIONS TO IT
\usepackage{caption} % DO NOT CHANGE THIS AND DO NOT ADD ANY OPTIONS TO IT
\usepackage{algorithm}
\usepackage{algorithmic}

\usepackage{newfloat}
\usepackage{listings}
\DeclareCaptionStyle{ruled}{labelfont=normalfont,labelsep=colon,strut=off} % DO NOT CHANGE THIS
\floatstyle{ruled}
\newfloat{listing}{tb}{lst}{}
\floatname{listing}{Listing}
\usepackage{color, soul}
\usepackage{url}

\begin{document}

\title{AI Risk Profiles: A Standards Proposal for Pre-Deployment AI Risk Disclosures}
\author {
    % Authors
    Eli Sherman \& Ian W. Eisenberg
}
\affiliations {
    % Affiliations
    Credo AI\\
    \{esherman, ian\}@credo.ai
}

\maketitle

\begin{abstract}
    As AI systems’ sophistication and proliferation have increased, awareness of the risks has grown proportionally \cite{nyt-biden-announcement}. In response, calls have grown for stronger emphasis on disclosure and transparency in the AI industry \cite{ntia-accountability-rfc, openai-governance-blogpost}, with proposals ranging from standardizing use of technical disclosures, like model cards \cite{mitchell2019model}, to yet-unspecified licensing regimes \cite{biz-insider-altman-licensing}. Since the AI value chain is complicated, with actors representing various expertise, perspectives, and values, it is crucial that consumers of a transparency disclosure be able to understand the risks of the AI system the disclosure concerns. In this paper we propose a risk profiling standard which can guide downstream decision-making, including triaging further risk assessment, informing procurement and deployment, and directing regulatory frameworks. The standard is built on our proposed taxonomy of AI risks, which reflects a high-level categorization of the wide variety of risks proposed in the literature. We outline the myriad data sources needed to construct informative Risk Profiles and propose a template-based methodology for collating risk information into a standard, yet flexible, structure. We apply this methodology to a number of prominent AI systems using publicly available information. To conclude, we discuss design decisions for the profiles and future work.
\end{abstract}

\section{Introduction}
While improvement in AI capabilities \cite{openai2023gpt, anthropic2023claude2} creates significant opportunities for societal benefit, AI systems pose numerous risks and potential harms. These risks take many forms, including social risks \cite{solaiman2023evaluating}, disruption to labor markets \cite{eloundou2023gpts}, creativity and culture \cite{franceschelli2023creativity}, democratic institutions \cite{jungherr2023artificial}, and even extreme risks \cite{shevlane2023model}. It is therefore critical that society have principled approaches to making decisions about AI deployment, which are founded on accurate, practical, and clear assessments of risk.

The range of stakeholders tasked with AI deployment decision-making is incredibly diverse. We argue that risk is an ideal level of abstraction that serves actors along the value chain. Non-technical decision-makers, for instance, readily consider the risk of adverse events when performing cost-benefit analyses \cite{vspavckova2015cost}. Regulators have found risk-framing similarly useful; the draft EU AI Act \cite{EUAIA} defines the obligations of AI developers in relation to risk. Risk also fits nicely into technical frameworks, corresponding to the expected value of possible adverse outcomes. In essence, risk can serve as a \emph{lingua franca} for AI system assessment.

Recent work has sought to clarify how risk evaluation and disclosure could serve this diversity of decision-makers.  \citet{shevlane2023model}, for instance, outlined a “transparency layer” within the AI development stack. This proposal makes the key observation that decision-makers will benefit from a standardized framework for pre-deployment evaluation. Such a convention should clearly outline risks in a unified taxonomy and be flexible enough to enable information distillation at several levels of technical depth.

Unfortunately, current conventions for AI risk disclosure do not satisfy these diverse needs. One popular disclosure artifact, Model Cards \cite{mitchell2019model}, aims to communicate the intended uses of a system, technical details, and ethical considerations to relatively technical audiences. However, Model Cards don’t standardize risk disclosures and are less relevant for multi-purpose AI systems, where expected but \emph{unintended} uses are critical concerns. A second approach is releasing detailed technical reports along with powerful AI systems, as pioneered by OpenAI \cite{openai2023gpt} and Anthropic\footnote {\url{https://www.anthropic.com/research}}. While highlighting modeling decisions and evaluations, this approach does not adequately support non-technical stakeholders, and could be improved to better serve technical consumers by expanding discussions of data sources, safety evaluations, and third-party assessments. Both approaches lack a standard risk taxonomy and fail to emphasize risk in characterizing AI systems.

\begin{figure*}
\centering
\includegraphics[scale=.3]{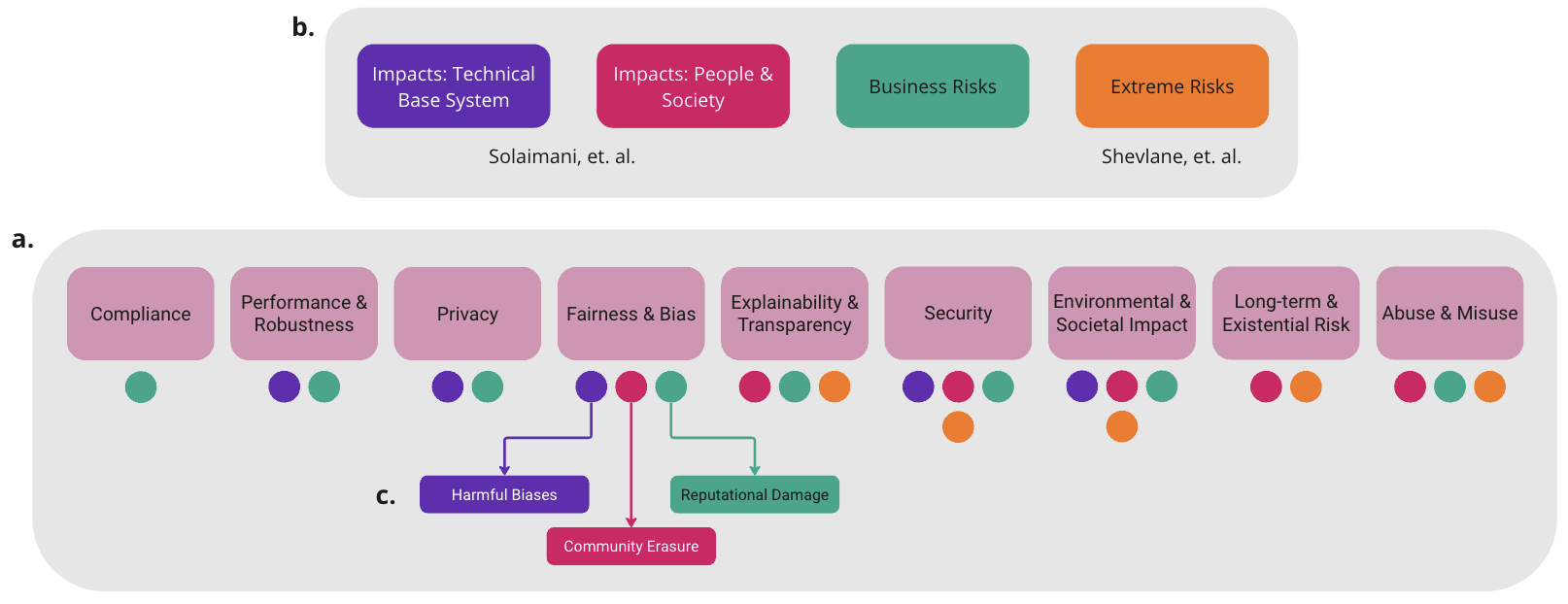}
\caption{Illustration of how the Risk Taxonomy (a) subsumes other risk categorization frameworks (b). Importantly, the Risk Taxonomy is expressive enough to capture multiple concerns, from corporate business interests in privacy and security to societal harms. Multiple risks exist under each category (c), and while direct 1-1 mapping is not always possible (e.g., many security risks are also privacy or societal risks), the taxonomy's primary goal is creating a standardized, high-level schema for risk identification and communication.}
\label{fig:risk-taxonomy}
\end{figure*}

In this paper, we propose a \emph{risk-centric} approach for assessing AI systems and a standardized reporting paradigm, which we call \emph{Risk Profiles}, for communicating assessment results. Inspired by other work on risk categories \cite{solaiman2023evaluating, shevlane2023model, barrett2023NTIA, EUAIA}, we propose a high-level taxonomy of risks which can guide risk assessment and disclosure. We establish a template for reporting the risks posed by an AI system and the mitigation measures provided by the system’s developer, and outline a methodology for collecting and distilling system information to generate Risk Profiles. We then apply this methodology to create Risk Profiles on several popular AI foundation models using publicly available information. We conclude with a discussion of alternative design decisions, how our risk profiling methodology can be enhanced with assistance from AI, and how Risk Profiles can help establish regulatory and industry standards for disclosure.

\section{A High-level Risk Taxonomy}
\label{sec:risk-set}
In this section, we detail our proposed taxonomy of AI risk categories. We center our taxonomy on high-level risk categories that subsume known risks, as exhibited in Fig. \ref{fig:risk-taxonomy}. This structure will guide a top-down approach to risk discovery and assessment, ensuring that no class of risks (e.g., social risks \citet{solaiman2023evaluating}) is ignored. Furthermore, the taxonomy is flexible, allowing evaluators to emphasize particular risk categories according to their needs. Additionally, the taxonomy can be used to anticipate both user-oriented harms, corporate incentives, and societal externalities, which is critical since many risk \emph{scenarios} reflect multiple \emph{risks} simultaneously – e.g., fairness, compliance, and societal impact are all present in the case of a discriminatory credit prediction system. Finally, while our taxonomy follows the inspiration of recent papers in using generative and general purpose systems (genAI) as the frame of reference, the risks posed by genAI systems are a superset of those pose by other AI systems and so our taxonomy can be effectively applied to both settings.

\textbf{Abuse \& Misuse:}\,\,
The potential for AI systems to be used maliciously or irresponsibly, including for creating deepfakes, automated cyber attacks, or invasive surveillance systems. Abuse specifically denotes intentional use of AI for harmful purposes.

\textbf{Compliance:}\,\,
The potential for AI systems to violate  laws, regulations, and ethical guidelines (including copyrights). Non-compliance can lead to legal penalties, reputation damage, and loss of trust. While other risks in our taxonomy apply to system developers, users, and broader society, this risk is generally restricted to the former two groups.

\textbf{Environmental \& Societal Impact:}\,\,
Concerns the broader changes AI might induce in society, such as labor displacement, mental health impacts, or the implications of manipulative technologies like deepfakes. It also includes the environmental implications of AI, particularly the strain on natural resources and carbon emissions caused by training complex AI models, balanced against the potential for AI to help mitigate environmental issues.

\begin{figure*}
    \centering
    \minipage{0.31\textwidth}
      \includegraphics[width=\linewidth]{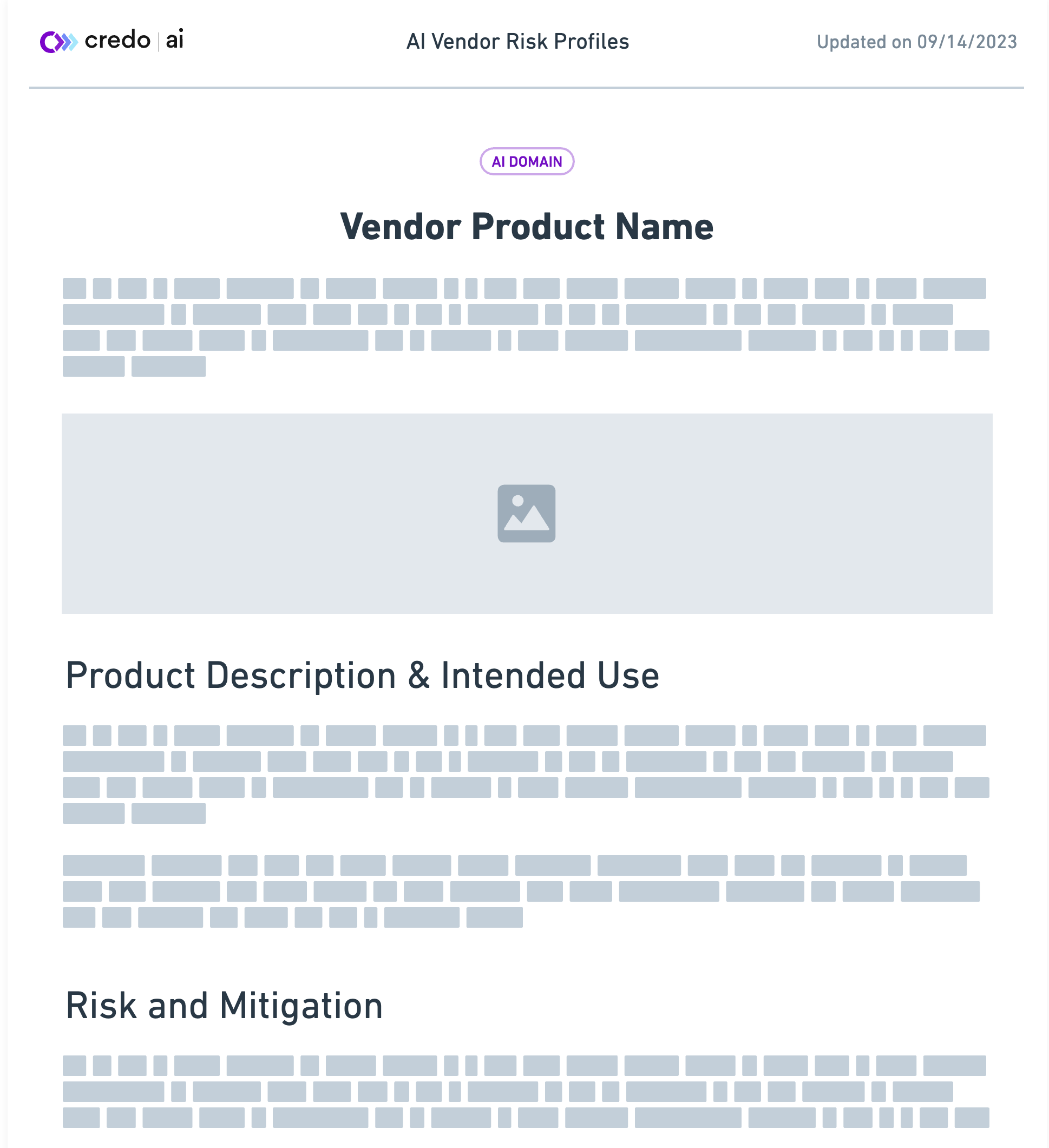}
    \endminipage\hfill
    \minipage{0.31\textwidth}
      \includegraphics[width=\linewidth]{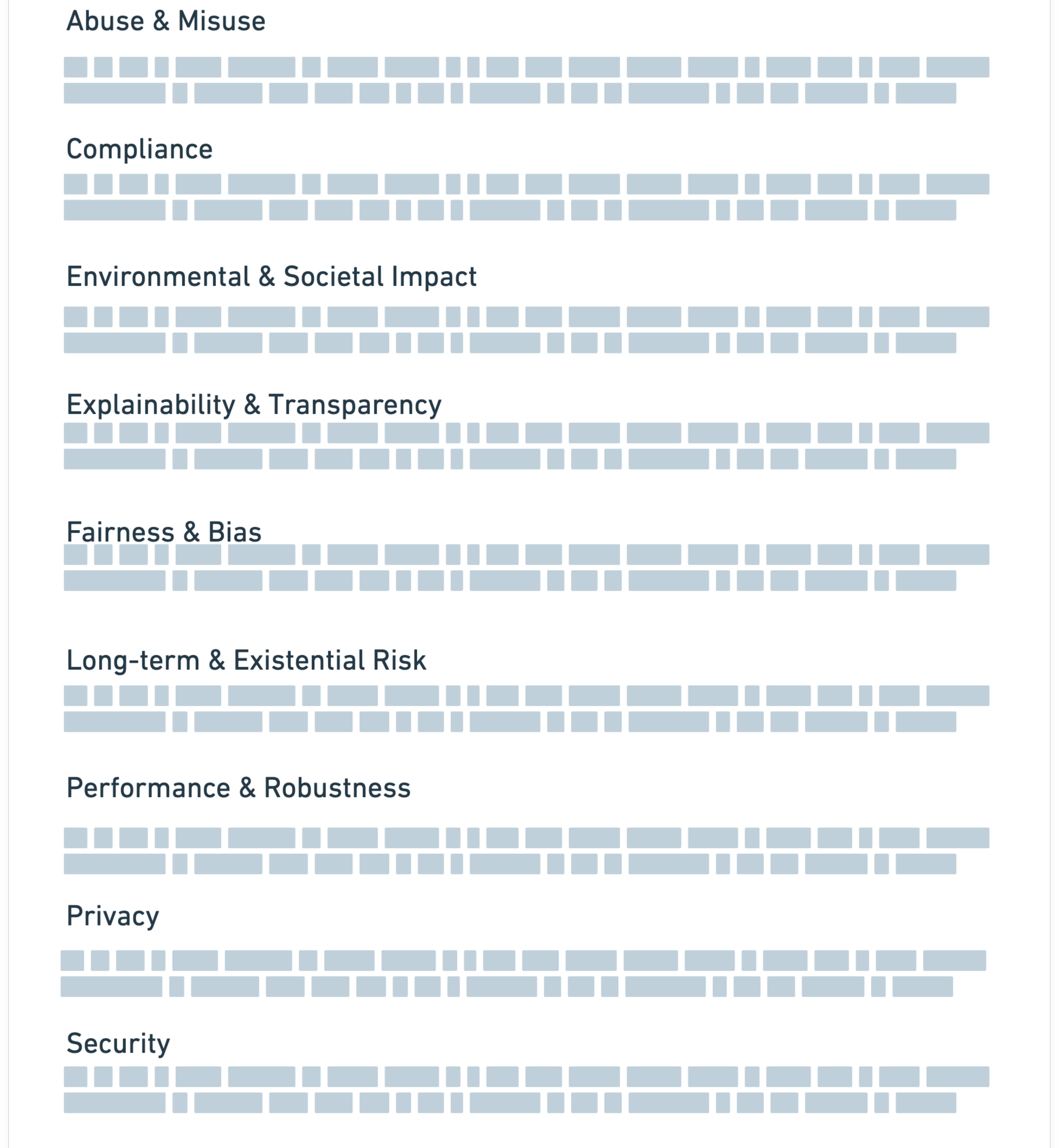}
    \endminipage\hfill
    \minipage{0.31\textwidth}
      \includegraphics[width=\linewidth]{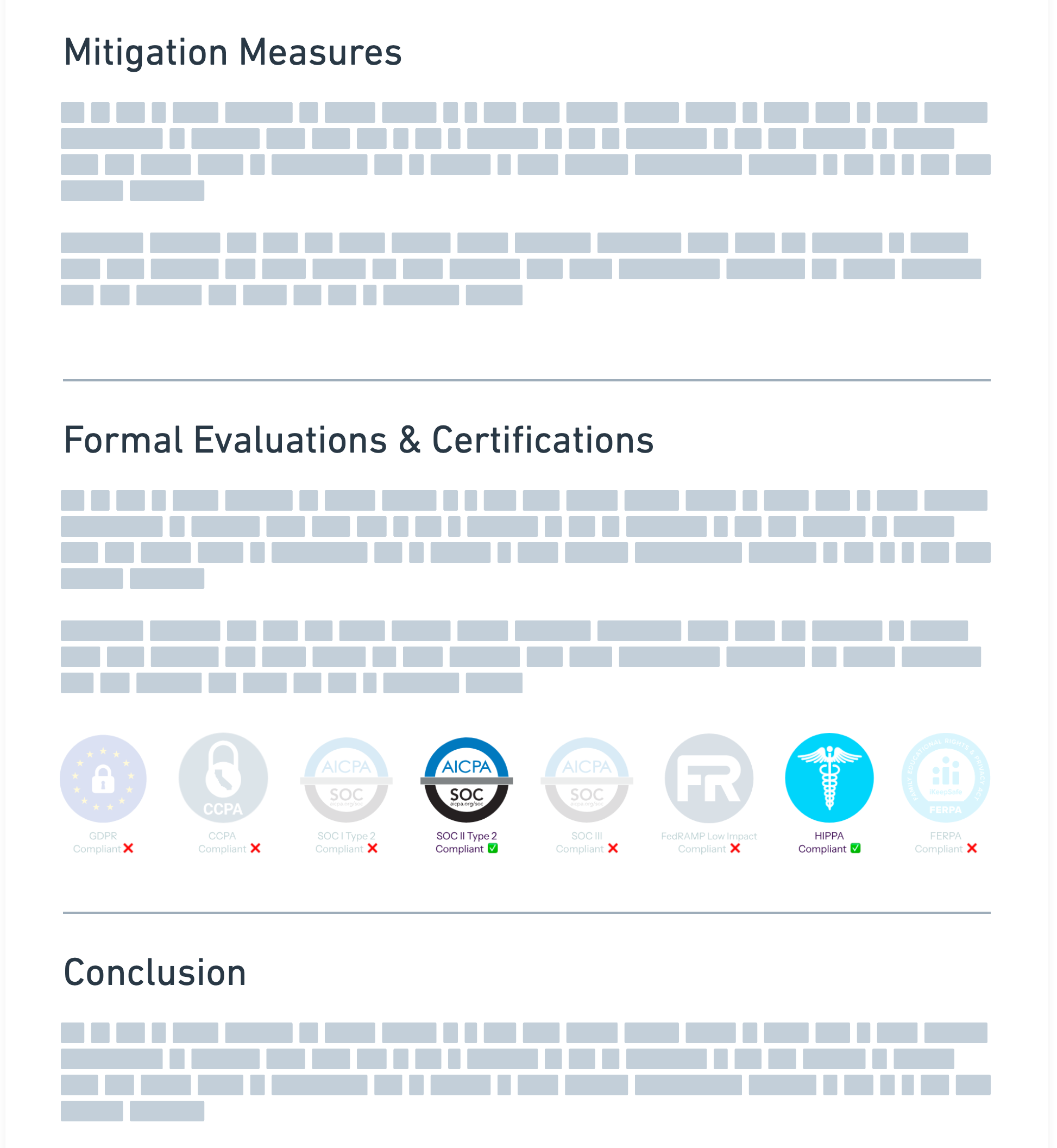}
    \endminipage
    \caption{Our proposed reporting template which supplements risk assessment with information about mitigations, evaluations, and certifications \& compliance.}
    \label{fig:report-template}
\end{figure*}

\textbf{Explainability \& Transparency:}\,\,
The feasibility of understanding and interpreting an AI system's decisions and actions, and the openness of the developer about the data used, algorithms employed, and decisions made. Lack of these elements can create risks of misuse, misinterpretation, and lack of accountability.

\textbf{Fairness \& Bias:}\,\,
The potential for AI systems to make decisions that systematically disadvantage certain groups or individuals. Bias can stem from training data, algorithmic design, or deployment practices, leading to unfair outcomes and possible legal ramifications.

\textbf{Long-term \& Existential Risk:}\,\,
The speculative potential for future advanced AI systems to harm human civilization, either through misuse or due to challenges in aligning AI objectives with human values.

\textbf{Performance \& Robustness:}\,\,
The AI's ability to fulfill its intended purpose accurately and its resilience to perturbations, unusual inputs, or adverse situations. Failures of performance are fundamental to the AI system performing its function. Failures of robustness can lead to severe consequences, especially in critical applications.

\textbf{Privacy:}\,\,
The potential for the AI system to infringe upon individuals' rights to privacy, through the data it collects, how it processes that data, or the conclusions it draws.

\textbf{Security:}\,\,
Encompasses potential vulnerabilities in AI systems that could compromise their integrity, availability, or confidentiality. Security breaches could result in significant harm, from incorrect decision-making to data leakage.

\section{A New Risk Disclosure Paradigm}
Here we detail our methodology for generating AI Risk Profiles. For illustration, we compiled profiles that exemplify our risk taxonomy and report template for five popular AI systems: Anthropic’s Claude, OpenAI’s GPT APIs, Microsoft Copilot, GitHub Copilot, and Midjourney. These can be found at \url{https://credo.ai/ai-vendor-directory}.

\begin{figure*}
    \centering
    \includegraphics[scale=.6]{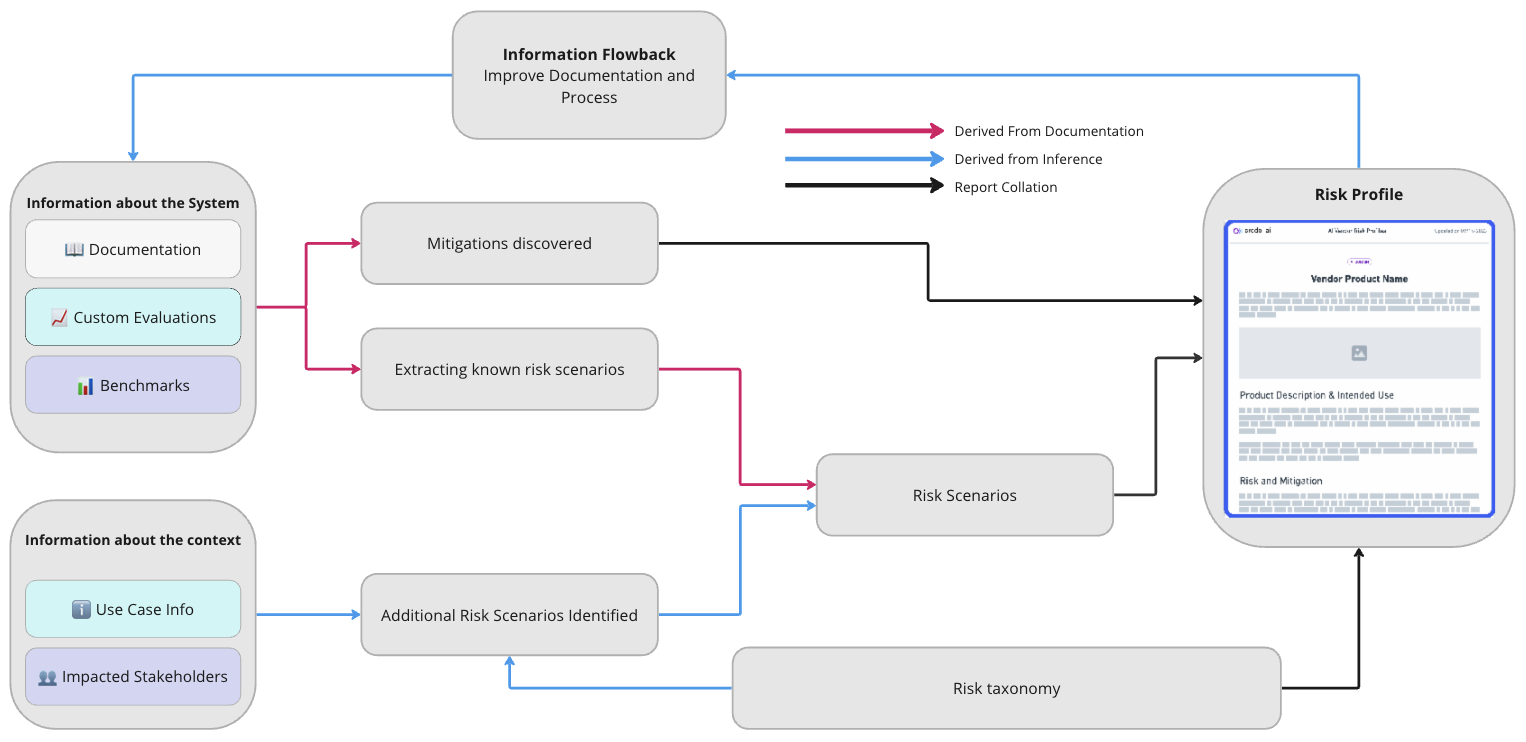}
    \caption{Our methodology for synthesizing available system information into risk profiles. Arrows represent flows of information, either via inference or directly from documentation. Risk insights distilled from the profile creation process should flow back to improve base documentation.}
    \label{fig:profile-methodology}
\end{figure*}

\subsection{Identifying Risk Scenarios}
To apply the risk taxonomy to a particular AI use case, report creators must identify applicable \emph{risk scenarios}, which cover adverse outcomes that arise from any use of the system (not simply intended use). The risk profiler must assess the AI system against a set of possible scenarios and infer whether each scenario applies based on the implicit probability of the scenario and the magnitude of impact. At present, risk scenario discovery relies on subject-matter expertise with respect to AI’s potential harms, with significant latitude for subjectivity. We expect that as the AI risk assessment field matures, practitioners will better define common risk scenarios, develop tools to support risk-scenario discovery, and establish conventions for determining the thresholds of relevance for each scenario.

It is critical that the identified risk scenarios are relevant to the Risk Profile’s target audience. Profiles compiled by independent third parties can approximate the risk scenarios relevant to a downstream user of the system, but comprehensive profiling requires incorporating all of the expertise and information that exists for the system.

\subsection{Surfacing Risk-Relevant Information}
In practice, risk scenario identification relies on inference from existing AI system documentation, which can be drawn from numerous sources including: 
\begin{itemize}
\item academic articles which detail system design decisions, training data composition, fitting procedures, and evaluations of system behaviors;
\item marketing materials about the system or the developer’s general research and design practices; 
\item independent evaluations of the system, including both replication studies and use case-specific evaluations; and
\item evaluations of analogous systems.
\end{itemize}
The availability of each source will depend on the report creator’s relationship with the system developer, as many developers are reluctant to disclose design and training information under pretenses of competition or potential for misuse. This disclosure paradigm will bring clarity to the risk evaluation process by highlighting \emph{where} risk information comes from and the causes of information gaps.

\subsection{Report Creation}
Figure \ref{fig:report-template} illustrates our proposed template for AI risk and mitigation reporting. The key sections include an introduction, an analysis of risks and mitigations, a discussion of formal evaluations of the system, and enumeration of which laws, regulations, and standards the system conforms to. We wrote our reports independently, without partnering with the AI vendors, and so our Risk Profiles represent a particular instantiation of this template. The general methodology for profile creation is summarized by Fig. \ref{fig:profile-methodology}.

Our instantiation of a Risk Profile begins with a short executive summary which highlights the top-line risks users and deployers should be aware of. The introduction then summarizes the AI system, its inputs and outputs, and its intended uses according to the developer. We also comment on the developer’s reputation, making sure to highlight the role of any key outside stakeholders, like investors.

\begin{figure*}
    \centering
    \includegraphics[scale=.4]{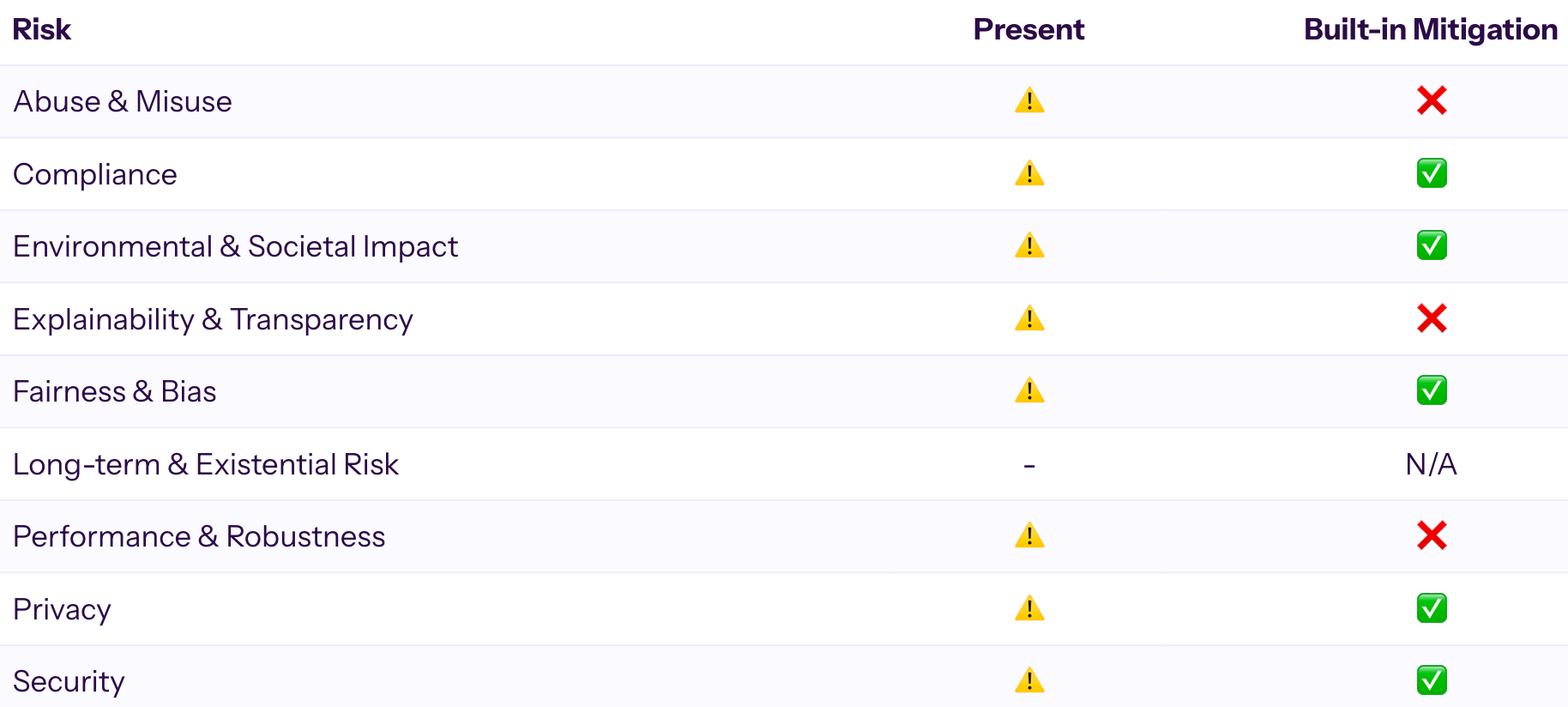}
    \caption{A summary of the risks present in GitHub Copilot and accounting of which risks GitHub has built mitigations for.}
    \label{fig:risk-summary}
\end{figure*}

The Risk and Mitigation analysis serves as the central section of the profile. We first provide a risk and mitigation summary table, as in Fig. \ref{fig:risk-summary}. The table identifies which of the 9 risks from our taxonomy are relevant to the system and whether the developer provides mitigation measures to address those risks. We then performed the risk scenario identification process described above for each risk category. Given our target audience of corporate procurement decision-makers and AI-informed regulatory organizations, we adopted an inclusive mindset for this step. We surfaced risk scenarios that were, in our view, plausible and relevant to those audiences, emphasizing scenarios that have received attention in public discourse in recent months. Likewise, while we discussed all developer-provided mitigations, we were cautious; given the difficulty of evaluating the efficacy of most mitigations, we stressed that risk \emph{reduction} should not be confused with \emph{elimination}.

The evaluations section focuses on quantitative benchmarks, relying primarily on the developers’ published evaluations. Where possible we highlight academic benchmarks (e.g. MMLU for large language models), but cite closed evaluations for dimensions of model behavior where no public evaluation exists. We also cite third-party qualitative metrics, such as chatbot rankings calculated by LMSys \cite{zheng2023judging}. We expect this section to expand to include further model-specific and qualitative evaluations (e.g., ARC’s evaluation of GPT4, \cite{openai2023gpt}), as evaluation methods mature \cite{perez2022discovering}.

We check system compliance with 8 standards and regulations, including the EU’s General Data Privacy Regulation, California’s Consumer Privacy Act, and SOC II. These are among the most relevant standards and laws for enterprises.

\section{Discussion}
\subsection{Design Decisions}
Generating our five Risk Profiles for popular AI systems entailed making several design decisions informed by our target audience and our independent position. These decisions point to axes of variation where report creators can make their profiles fit for purpose.

\textbf{Profiler Identity}\,\,
Perhaps the most influential factor in Risk Profile compilation is the relationship between the Risk Profile creator and the AI system developer. Our example profiles reflect our independent synthesis of public information about each system. This avoids some biases, such as omission of reputationally-damaging information about the developer, that would arise if the profiles were first-party generated. For instance, it is particularly difficult for a developer to impartially assess their system’s `Explainability and Transparency’ risks. On the other hand, developer involvement has benefits: they are likely knowledgeable about system-specific risk scenarios and privy to additional evaluation details that make a profile comprehensive.

\textbf{Risk and Mitigation Summary}\,\,
As a multi-page report, the Risk Profiles benefit from summary information – such as executive summaries and risk “quantification” like the table in Fig. \ref{fig:risk-summary} -- which enables readers to efficiently glean the most important details in the transparency report. While we presented a simple binarized view of risks and mitigations, there are numerous alternatives to aggregating and quantifying risk-related information. For instance, one could provide counts of which risks (scenarios) apply to the system – either in absolute terms or as a fraction of the total number of risks considered. Further sophistication is possible by positing system-level or per-risk  scores  which account for the probability of the risk (scenario) and the impact were that risk to be realized, or analogously the degree of mitigation achieved through a particular strategy.

All of these approaches, including our binarized presentation, have limitations. Binarizing and counting implicitly weigh all risks equally in terms of both probability and impact. This simplification could mislead report consumers. For instance, a distant, low-likelihood risk like a large language model-based (LLM) AI developing resource acquisition capabilities may be over-weighed relative to an immediate, high-likelihood risk like that same LLM encoding racial stereotypes into employment-related decisions. Risk scoring is also problematic. Absent actual adverse outcomes data, scoring relies on subjective weighting of competing risks, meaning the risk summarization will over-index to the risk profiler’s biases. We opted for the former option because it is more conservative. Nevertheless, there is clearly space to further develop these approaches – we envision a future scoring method which serves as a middle-ground by enabling transparent application of subjective weightings. This is an area of ongoing work.

\textbf{Evaluations}\,\,
The nascent state of AI evaluations, particularly for genAI systems, makes profiling challenging. Model developers are beginning to establish conventions around which benchmarks are relevant to each model type (e.g. MMLU \cite{hendrycks2020measuring} for LLMs) but more work is necessary. First, performance on traditional benchmarks is increasingly saturated \cite{maslej2023ai}, requiring the development of more challenging and comprehensive benchmarks. Second, current benchmark approaches are not comprehensive due to the difficulty in creating them (though automated evaluations could help solve this \cite{perez2022discovering}). Given the limitations of current benchmarks, profilers should supplement benchmarks with bespoke and qualitative evaluations like the results of red-teaming \cite{perez2022red}, alignment-oriented evaluations \cite{perez2022discovering}, and contextualized evaluations on use-case relevant datasets. Lastly, existing evaluation approaches poorly anticipate downstream impacts on stakeholders and society writ-large. Recent proposals for `soft’ deployments, such as `regulatory sandboxes’ \cite{truby2022sandbox}, could prove useful for improving impact-forecasting, as could explicit measures of societal impact \cite{solaiman2023evaluating}, both of which can be incorporated into Risk Profiles as supporting evaluation information.

Regardless of which evaluations are cited in the Risk Profile, the profiler has the responsibility to discuss their relevance and interpretation. Readers should be able to understand the relationship between each metric and the system’s  suitability for the deployment setting.

\subsection{AI-Assisted Profiling}
As AI systems get more sophisticated, both the volume of technical documentation and the size of the potential user base will increase. This will make it more difficult to scalably distill risk information for each stakeholder. While expert profilers are expected to remain critical for some time, we believe one of the most fertile areas for continued development of our profiling proposal is to leverage AI to aid in the research and report creation processes. As we have articulated, there are many areas where value judgments are critical during the profiling process, but there is ample opportunity for AI-systems to support by aggregating, distilling, and transforming risk-relevant information. As LLMs become more robust to hallucinations, it may be possible to make this process accessible directly to report consumers, who can specify their information needs and receive accurate risk assessments without expert input.

\subsection{Implications for Mandated Reporting Requirements}
Increasing public awareness of AI’s risks has led to calls for mandated risk disclosures by AI developers \cite{EUAIA, biz-insider-altman-licensing}. Our Risk Profile template and methodology address many of the concerns that motivate these proposals. For instance, the draft EU AI Act lays out transparency requirements for high risk AI systems and foundation models. Obligations cover details like a `Description of the capabilities and limitations of the foundation model’ or `Description of the model’s performance, including on public benchmarks or state of the art industry benchmarks’. While capabilities and performance are partially covered by artifacts like Model Cards, the Risk Profile expands this artifact with a detailing of mitigations, a comprehensive articulation of limitations, and the added benefit that our methodology provides a guide for \emph{contextualizing} the required risk information.
Two recent related research efforts – Algorithmic impact assessments and regulatory scorecards – point to immediate steps we can take to strengthen our proposals in service of a convention for regulation-mandated risk reporting. Algorithmic impact assessments \cite{reisman2018algorithmic} were conceived as a tool to inform regulators on algorithmic harms. In particular, they establish a principled process for identifying and evaluating likely harms to stakeholders and externalities. Our approach can incorporate this process to harden our evaluations section and fill gaps left by the present inadequacy of technical evaluations. Regulatory scorecards, like the one studied in \citet{bommasani2023eu-ai-act} for the EU AI Act, represent an alternative approach to risk summarization. They amount to a binary checklist, not of risk application, but of satisfaction of legal requirements. When a Risk Profile is explicitly compiled to show conformity with a specific regulation, adopting a scorecard can be hugely beneficial for clarifying and communicating how the AI system matches up against the requirements.

\subsection{Monitoring \& Ongoing Decision-Making}
Ultimately, the full benefits of Risk Profiles will be realized when they can inform \emph{ongoing} decision-making, continuing after  the initial deployment. As a system sees in-production use, the pre-deployment Risk Profile can be used as a guide for governance of the system. It can indicate the need for additional evaluations, further mitigation measures, or even changes to the system itself, like model re-training. It also can inform a montioring strategy which can lead to continuous updating of the Risk Profile, making it a living document. One resource that can support this ongoing risk assessment and risk management is AI incident databases \cite{mcgregor2021preventing}. Incident databases can provide crucial information which validates prior assessments or, at worst, points to unanticipated risk areas which were not identified in the initial Risk Profile. We intend to incorporate AI incident information in future iterations of our profiling methodology.

\bibliography{references}

\end{document}